\documentclass[10pt, a4paper]{article}
\usepackage{lrec2016}
\usepackage{multibib}
\newcites{languageresource}{Language Resources}
\usepackage{graphicx}
\usepackage{amsmath,amssymb,amsfonts}
\usepackage{algorithmic}
\usepackage{graphicx}
\usepackage{textcomp}
\usepackage{hyperref}
\usepackage{cleveref}[2012/02/15]
\usepackage[table]{xcolor}
\usepackage{nicefrac}
\usepackage{siunitx}
\usepackage{array}
\usepackage{float}
\usepackage{stfloats}
\usepackage{array,booktabs}
\newcolumntype{M}[1]{>{\centering\arraybackslash}m{#1}}
\usepackage{xcolor,colortbl}
\def\BibTeX{{\rm B\kern-.05em{\sc i\kern-.025em b}\kern-.08em
    T\kern-.1667em\lower.7ex\hbox{E}\kern-.125emX}}

\crefformat{footnote}{#2\footnotemark[#1]#3}

\makeatletter
\newcommand{\thickhline}{%
    \noalign {\ifnum 0=`}\fi \hrule height 1pt
    \futurelet \reserved@a \@xhline
}

\usepackage{xcolor}

\usepackage{epstopdf}
\usepackage[latin1]{inputenc}

\title{Studying Socially Unacceptable Discourse Classification (SUD) through different eyes: \emph{"Are we on the same page ? "}  \\ }

\name{Bruno Machado Carneiro, Michele Linardi, Julien Longhi}

\address{ENSEA Engineering School,  ETIS UMR-8051 CY Cergy Paris Universit\'e, AGORA CY Cergy Paris Universit\'e  \\
bruno.machadocarneiro@ensea.fr, \{michele.linardi, julien.longhi\}@cyu.fr\\}

\abstract{
We study Socially Unacceptable Discourse (SUD) characterization and detection in online text.
We first build and present a novel corpus that contains a large variety of manually annotated texts from different online sources used so far in state-of-the-art Machine learning (ML) SUD detection solutions.
This global context allows us to test the generalization ability of SUD classifiers that acquire knowledge around the same SUD categories, but from different contexts.
From this perspective, we can analyze how (possibly) different annotation modalities influence SUD learning by discussing open challenges and open research directions.
We also provide several data insights which can support domain experts in the annotation task.
\textit{Accepted for publication in the International Conference on CMC and Social Media Corpora for the Humanities (University of Mannheim, Germany, 2023)}. 
\newline 
 }

\begin{document}

\maketitleabstract

\section*{Acknowledgment}
The work presented in this paper is part of the ARENAS project. 
This project has received funding from the European Union's Horizon Europe research and innovation programme under grant agreement No:101094731.

\section{Introduction}

During these last two decades, the massive popularisation of social media has been changing the way people communicate, interact and collect worldwide news. 
The dissemination speed rate and the possibility to quickly reach a large audience are some clear advantages of modern social network platforms.
By contrast, the potential anonymity and sense of impunity can bring out the worst in people and made them sharing ideas that would not be socially acceptable otherwise.
Socially Unacceptable Discourse~\cite{Sulc2020NoRF} (SUD) typically occur in various form;
%1 hate offensive, %2 radicalization ideologies
The use of offensive and abusive language represent a common form of SUD, but it is also important to note that controversial narratives are not necessarily bad or immoral, but they closely relate to radicalization and ideologies.
Clear contexts in the recent history are the Covid-19 crisis and the the Russian invasion of Ukraine.
During these periods, we have witnessed several cases of public debate radicalization, especially favored by the circulation of distorted information~\cite{vaccines10030481} that jeopardizes the knowledge acquisition of complex systems and environments.

Another particular trait of SUD is the presence of distinctive grammatical characteristics.
To model these features, we require identifying several grammatical substructures such as residual representations, use of pronouns, and future tense~\cite{Ascone_Longhi_2018,Maiti2020GrammaticalFO}.

We note that, in publicly annotated corpus used so far by the Machine Learning community, no standard or common guidelines for SUD annotation exist~\cite{fiser-etal-2017-legal} despite the adoption of the same terminology and/or tags.
It derives that different SUD definitions may potentially share overlapping characteristics, or on the other hand a single category may cover text instances with divergent features depending on the context.
Furthermore, annotators bias can also play a decisive role as reported by previous works~\cite{bias1,yuan_transfer,racial_bias}.

In this scenario, it is reasonable to expect a poor generalization capability of ML SUD classifiers trained in a specific context~\cite{yuan_unseen}. 
To that extent, we study and evaluate the capability of current state-of-the-art Deep Learning models to characterize SUD on different grounds. 

Other works have recently considered the zero-shot learning problem in hate speech detection, where transfer learning is tested and measured on binary (hate/no hate)~\cite{toraman2022large} and on multi-class~\cite{yuan_unseen} classification.
In this context, we sketch and propose a different approach that first aims to test transfer learning at a class level rather than a dataset level.
This approach permits us to provide more interpretable insights on the SUD semantic and to test the transfer over different annotation guidelines on the same speech categories.

%Our objective was to made a step towards a more general understanding of SUD, by unifying tasks that are most of the time treated separately in the literature. Our hope is that a model with a general understanding of SUD could be used in a future work for Transfer Learning into a extremist narrative's detection model. 

%In our experiments, we trained BERT to classify between all the different types of SUD occurring in a unique corpus built by the concatenation of available datasets in the literature.

%We also study the embedded generated by the model analysing the relationship between the different classes. 
%Additionally, we propose a novel binary SUD classification, which generalizes SUD categories in a single class. 
%The proposed model achieves a Macro F1 score of 88.5\% on our testing set.
%Furthermore, we compare it with models trained on each available dataset.
%The proposed classifier over-performs the state-of-the-art results in several datasets, while maintaining similar performance to the competitors in the others.

%In summary, we perform classification between the 12 types of SUD present on our corpus, over-performing the state-of-the-art in 4 of the 13 used datasets. We also present a generic SUD detector, achieving a Macro F1 score of 88.5\%. Furthermore, we studied the embedding generated by our trained models, using dimensionality reduction to generate 2 dimensional plots.

%\commentRed{Quickly introduce (summarize) the results we add in the experiments. A recapitulating list usually works fine}

\section{Socially Unacceptable Discourse Corpora}

\begin{figure*}[tb]
  \centering
  \includegraphics[trim={1.5cm 8cm 0cm 4cm}, scale=0.48 ]{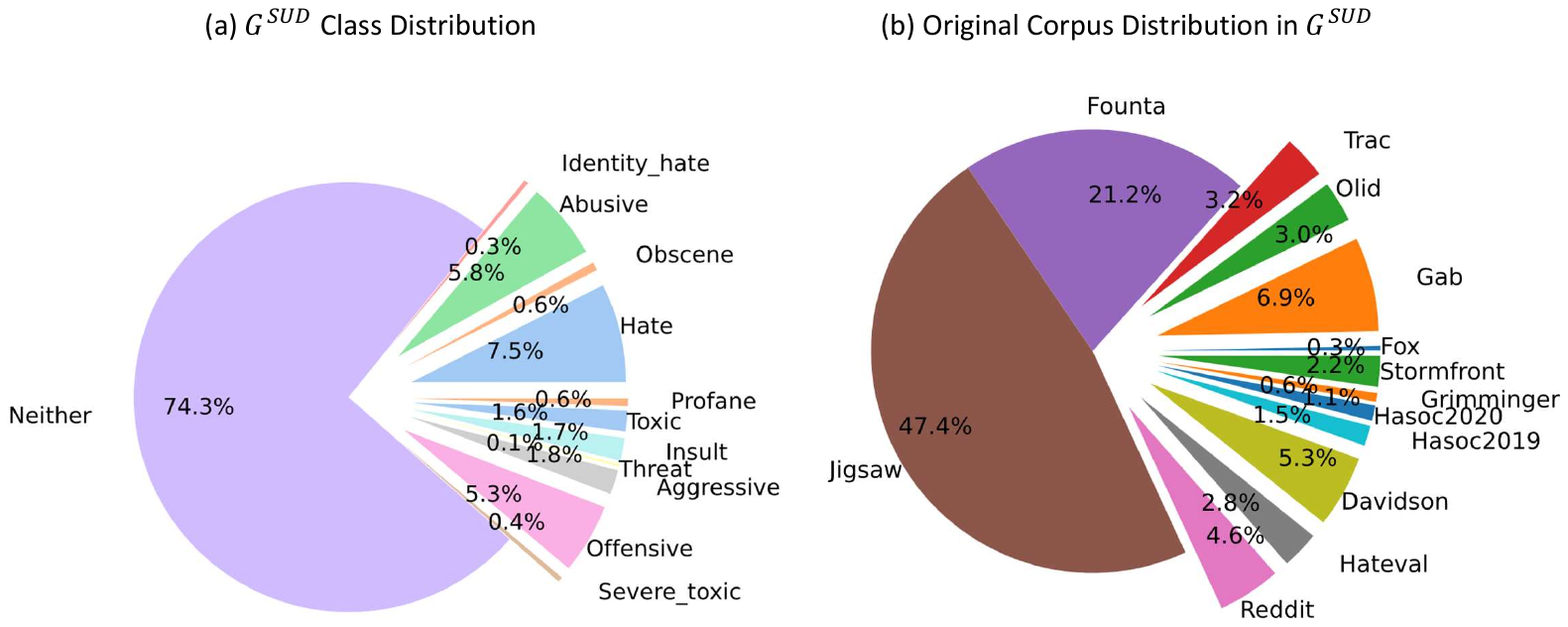}
  \caption{ (a) $G^{SUD}$ Class distribution, (b) Corpus distribution in $G^{SUD}$}
  \label{fig:class_dist}
\end{figure*}

We report the corpora we consider in our study in table~\ref{tab_all_corpus}.
%These datasets form a novel and extensive ground, where to evaluate current state-of-the-art SUD detection capability.
We use data from various sources recently adopted to assess the performance of state-of-the-art ML solutions for automatic SUD detection (e.g., hate speech detection, sentiment, toxicity, radicalization, and ideology analysis).

%In section \ref{datasets} we provide further details about the corpus we built.

%In section \ref{model} we discuss the NLP model choice, along with adopted data processing. Our main goal is to characterize SUD and evaluate the resulting model in a transfer learning context. 
%A research question we want to address concerns the ability of state-of-the-art Machine Learning models to generalize over new data in which either new labels exist (unknown classes) or well know SUD types connect to different patterns.
%Therefore, we focus more on the embedding generation task rather than the absolute classification results. Details about the training are given in section~\ref{training}.

%\subsection{Datasets}\label{datasets}

\begin{table*}[tb]
\resizebox{17.5cm}{!}
{
\begin{tabular}{ | c | c | c | c | c | c | }
 \hline
 \textbf{Dataset} & \textbf{Sample type} & \textbf{\# Samples} & \textbf{Topic} &   \textbf{ Best performing SUD classifier}  & \textbf{F1 Macro (\%)}  \\
 \hline
 Davidson~\cite{grimminger} & Tweets & 25,000 & Generic & BERT& 93 \\
 \hline
 Founta~\cite{swamy}  & Tweets & 100,000 & Generic & BERT & 69.6 \\
 \hline
 Fox~\cite{yuan_unseen}  & Threads & 1,528 & Fox News Posts & BERT & 65 \\
 \hline
 Gab~\cite{reddit_gab}  & Posts & 34,000 & Generic & CNN & 89.6 \\
 \hline
 Grimminger~\cite{grimminger}  & Tweets & 3,000 &  US Presidential Election & BERT & 74 \\
 \hline
 HASOC2019~\cite{hasoc19_winner}  & Facebook, Twitter posts & 12,000 & Generic & LSTM + Attention & 78.8 \\
 \hline
 HASOC2020~\cite{hasoc2020_w}  & Facebook posts &  12,000 & Generic & XLM-RoBERTa & 90.3\\
 \hline
 Hateval~\cite{mSVM} & Tweets & 13,000 & Misogynist and Racist content & mSVM/BERT & 75.4 \\
 \hline
 Jigsaw~\cite{jigsaw_w}  & Wikipedia talk pages & 220,000 & Generic & Bi-GRU + Attention & 78.3 \\
 \hline
 Olid~\cite{olid}  & Tweets & 14,000 & Generic & CNN & 80\\
 \hline
 Reddit~\cite{yuan_unseen}  & Posts & 22000 & Toxic subjects & BERT & 85 \\
 \hline
 Stormfront~\cite{mSVM}  & Threads & 10,500 & White Supremacy Forum & BERT & 80.3 \\
 \hline
 Trac~\cite{trac_w} & Facebook posts & 15,000 & Generic & LSTM & 64 \\
 \hline
\end{tabular}
}
\caption{Best performing SUD classification model on each dataset. }
\label{tab_all_corpus}
\end{table*}

We selected $\mathbf{13}$ publicly available datasets containing $\mathbf{470,768}$ samples distributed over 12 classes. 

%Due to space limitations, we report a detailed description of their main characteristics and collecting methods in the appendix.

We generate a unique English text corpus by concatenating all the $13$ datasets, denoting it with the label $G^{SUD}$.
Note that the datasets we concatenate in $G^{SUD}$ share multiple overlapping SUD labels, which identify the same SUD category.
We consider the presence of bias and ambiguities as physiological, and identifying and analyse the concerned instances is under the lens of our research.

In figure~\ref{fig:class_dist}(a), we report the instances distribution over SUD classes.
Note that the \emph{neither} class subsumes all texts that do not fall in any SUD categorizations proposed by the annotators. 
As expected, SUD classes have a sensitive lower support compared to the \emph{neither} class denoting the typical class imbalance setting of the SUD detection problem.

Figure \ref{fig:class_dist}(b) illustrates the ratio of each dataset with respect to the global corpus. 
We observe that Jigsaw and Founta contain together more than $60\%$ of the data. 

\subsection{Datasets}
Here, we provide the details of each dataset we join in $G^{SUD}$.

\textbf{Davidson \cite{davidson}} contains around 25,000 tweets labelled as being hateful, offensive or neither of those randomly sampled from a set of 85.4 million tweets produced by 33,458 different users. 
Each sample was labelled by at least three different annotators.

\textbf{Founta \cite{founta}} contains about 100,000 tweets, labeled with four categories: abusive, hateful, normal, and spam. 
In this dataset, a variable number of users (between five and ten) have annotated each sample.

\textbf{Fox \cite{fox}} contains $1528$ comments posted on ten different popular threads on the Fox News website. 
In these data, two native English speakers have produced labels to differentiate hateful from normal content following the same annotation guidelines.

\textbf{Gab \cite{reddit_gab}} contains $34,000$ samples extracted form Gab, a social media, where users commonly share far-right ideologies~\cite{gab_info}, annotated in the Amazon Mechanical Turk\footnote{\label{note2}\url{https://www.mturk.com/}} platform, where at least 3 annotators provided a label for each sample. 

\textbf{Grimminger \cite{grimminger}} contains $3,000$ tweets on 2020 presidential election topic in the United States. The samples were labelled between hate speech or not by three undergraduate students, who discussed the annotation guidelines during the labelling process. 

\textbf{HASOC2019~\cite{hasoc2019}} and \textbf{HASOC2020~\cite{hasoc2020}} are datasets proposed in the Indo-European Languages (HASOC) challenge, which contain $12,000$ English text samples extracted from Twiter and Facebook labeled between hateful, offensive, profane or neither of those.

\textbf{Hateval \cite{hateval}} gathers around $13,000$ tweets containing hateful and normal speech. The hateful content originates from accounts of potential victims of misogynism and racism. 

\textbf{Jigsaw\footnote{\url{https://www.kaggle.com/c/jigsaw-toxic-comment-classification-challenge}}}~\cite{jigsaw_w} is a dataset provided in the Toxic Comment Classification Challenge. It contains about 220,000 samples extracted from Wikipedia talk pages differentiated into seven classes: toxic, severe toxic, obscene, threat, insult, identity hate, and neither of the previous.

\textbf{Olid \cite{olid}}  contains $14,000 $ tweets annotated using the Figure Eight Data Labelling platform~\footnote{\url{https://f8federal.com/}}.
In this context, tweet selection is executed by keyword filtering and human annotation.

\textbf{Reddit \cite{reddit_gab}} has 22,000 samples extracted from Reddit, labeled for hate speech detection by Amazon Mechanical Turk users. Before the labeling task, the text got selected according to a list of toxic subjects on the Reddit platform.

\textbf{Stormfront \cite{stormfront}} contains 10,500 samples taken from a white supremacy forum called Stormfront and divided into four classes: hate, no hate, related, and skip. 
The related class contains statements that can not be considered hateful without considering their context.
Text belonging to the skip class does not contain enough information to determine if it can be classified as hateful.

\textbf{Trac~\cite{trac}} dataset gathers $15,000$ Facebook posts and comments classified into aggressive and non-aggressive language.

\section{SUD Deep Learning Models } \label{model}

In this section, we introduce and describe the state-of-the-art Deep Learning models adopted for the SUD detection task in previous works.  
In Table~\ref{tab_all_corpus}, we show the best performer in each corpus.
Here, we report the Macro F1 score, which is the recommended averaging method for F1 score when dealing with class imbalance.
It is calculated by averaging the sum of the F1 score of each class. 

Recall that the F1 score reports the harmonic mean of precision and recall of a classification model.
For a particular input class, we compute the precision (P) and recall (R) of a SUD classifier as follows: $ P = \frac{TP}{TP + FP} $, and  $R = \frac{TP}{TP + FN}$, where TP denotes the number of correctly classified instances of the input class (true positive), FP denotes the number of occurrences that are wrongly assigned with the input class label (false positive), and FN represents the number the input class samples that are erroneously classified (false negative). 
Hence we have that $F1 = 2 \times \frac{P \times R}{P + R}$.

From Table~\ref{tab_all_corpus}, we observe that \textbf{BERT} (Bidirectional Encoder Representations from Transformers~\cite{bert}) is the best performer model in the majority of the datasets. 
BERT adopts a Deep Learning (DL) architecture released by the Google AI Language team in early 2019, which is pre-trained by masked language model (MLM) and next sentence prediction (NSP) tasks over a large corpus of English data containing more than 3B words~\cite{bert}.
MLM consists of training the model to predict masked tokens in the corpus sentences, whereas the NSP training aims to predict if two sentences form a sequence in the original text.
XLM-RoBERTa~\cite{ConneauKGCWGGOZ20} is a multilingual variant of the original BERT model.

BERT has clearly shown its superiority over other types of DL models previously adopted in SUD classification, such as Convolutional Neural Networks (CNN)~\cite{reddit_gab} and Long-short term memory networks LSTM~\cite{hasoc19_winner}. 
The attention mechanism used by BERT represents a robust solution that can better learn long-range token dependencies, avoiding the limitation of LSTM networks, which assumes that each token depends only on previous ones. 
By contrast, BERT learns relationships considering all the tokens in a  sentence simultaneously.

In this work, we evaluate the SUD classification performance of BERT in the heterogenous corpus we construct.
In the next section, we present all the research questions we address, discussing the results we obtain.
%We de BERT seems to dominate the state-of-the-art for SUD classification. Even for the datasets on which it doesn't has the best reported performance, it can still have competitive results. For example, on the Hateval dataset, Yuan and Rizoiu \cite{yuan_unseen} report a Macro F1 Score of 75.2\% using BERT, having technical draw with the mSVM approach presented by MacAvaney et al. \cite{mSVM}. On those grounds, we decided to use BERT to proceed with our study.

\section{Experiments} \label{experiments}

\begin{table}[htbp]
\resizebox{8cm}{!}
{
\begin{tabular}{ | c | c | c | c | }
\cline{2-4}
\multicolumn{1}{c|}{} & \multicolumn{3}{c|}{\textbf{F1 Score (\%)}} \\
\hline 
\textbf{Training set} & \textbf{Macro} & \textbf{Weighted} & \textbf{Micro} \\
 \hline
 $G^{SUD}$  & 53.9 & 86.8 & 87.1 \\
 \hline
 $G^{SUD}$ Balanced  & 51.3 & 85   & 84.5 \\  
 \hline
 $G^{SUD}$ with Neither Undersampled  & 58.5 & 73.7 & 73.9 \\  
 \hline
 $G^{SUD}$ balanced with Neither Undersampled   & 56.8 & 72.5 & 72.1 \\    
 \thickhline    
 $G^{SUD}$ (Binary classification) & 88.5 & 91.3 & 91.2 \\    
 \hline
 $G^{SUD}$ balanced (Binary classification)  & 89.7 & 89.7 & 89.7 \\   
 \hline
\end{tabular}
}
\caption{Comparison between all experiments}
\label{tab2}
\end{table}

\subsection{Multiclass SUD Classification} \label{Multiclass}

To conduct our experimental evaluation, we use the BERT\textsubscript{BASE}~\cite{bert,yuan_unseen} model pre-trained by WordPiece tokenizer algorithm.
For the sake of reproducibility, we provide the code and the data used in the experiments along with the relative instructions in an online repository~\cite{paperRepo}.

To perform SUD classification, we connect BERT pooled output layers to a Multi-Layer Perceptron (MLP) architecture that contains $12$ output neurons (one per class). %classification heads (one per SUD type).
We have fine-tuned the MLP layer of proposed model on the $G^{SUD}$ corpus using a $80\%/10\%/10\%$ splitting ratio for training, validation, and testing respectively. 
We have adopted a stratified sampling technique to keep the same class distribution throughout the three splits. 
Hyper-parameters have been tuned by performing several complete training rounds, picking the setting with the best validation performance.

The research questions we want to address are the following: \emph{Which are the state-of-the-art model generalization capability in a global context? What are the main challenges that hamper the SUD modelling effectiveness?}  

Table~\ref{tab2} contains the results, where we report Macro, Weighted and Micro F1 score of the SUD classification.
Note that the Weighted F1 weighs the global F1 average according each class support, whereas the Micro F1 score computes a global F1 making no distinction across classes.

Considering that $G^{SUD}$ contains highly unbalanced SUD classes, we repeat classification tasks after training our model on a balanced dataset.
To that extent, we have performed random oversampling of minority classes as suggested by several works~\cite{yuan_unseen,swamy,mSVM}.

Furthermore, given the dominance of the \textit{neither} class, we also consider a setting with under-sampled non-SUD text (neither class).
Here, we have selected $10\%$ of the non-SUD samples in a stratified way, maintaining the same proportion of the \textit{neither} class samples in every dataset. 

We note that undersampling the \textit{neither} class has a sensitive effect on the model prediction capability as the Macro F1 score increases.
On the other end, reducing the neutral class causes an increment of model errors for the \textit{neither} class (majority class) as we observe a significant reduction of the Weighted and Micro F1 scores. 
It follows that coping with such an imbalance between non-SUD and SUD samples represents a concrete challenge (typically occurring in a real-world scenario), which is amplified in the extended corpus under consideration. 
 
We also notice that producing a balanced class scenario by performing random oversampling does not provide any significant benefit.
This suggests that class imbalance is only a joint cause of the model discrimination capability. 

\begin{figure}[tb]
  \centering
  \includegraphics[trim={0cm 10cm 0cm 3cm}, scale=0.63]{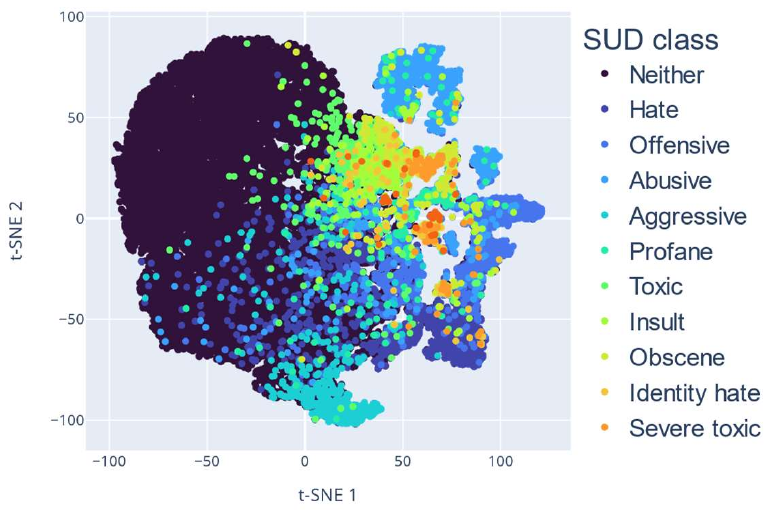}
  \caption{Two components t-SNE visualization of samples embedding produced by BERT output pooled layer.}
  \label{fig:t-SNE}
\end{figure}

To better understand how the adopted model discriminates SUD classes, we visualize the generated text representation (output of BERT output pooled layer).
To reduce the dimensionality of the latent space, we apply t-distributed Stochastic Neighbor Embedding (t-SNE). 
Figure~\ref{fig:t-SNE} shows the plot computed over the testing set, with a model trained on the complete corpus $G^{SUD}$.
%Due to space limitations, in the appendix, we report a table containing the classification performance for each class. 
In Table~\ref{tabPerClassPerDataset} we report the Macro F1 score of SUD classification in $G^{SUD}$ for each dataset and each class.
Note that each line in this table corresponds to a different model, trained only on the specified dataset, while the first line is the result obtained using the model trained on $G^{SUD}$.

\begin{table*}[htbp]
\centering
\resizebox{17.5cm}{!}
{
    \begin{tabular}{ | c | c | c | c | c | c | c | c | c | c | c | c | c | }
     \cline{2-13}
     \multicolumn{1}{c|}{} & \multicolumn{12}{c|}{Macro F1 Score (\%)} \\
     \cline{2-13}
     \multicolumn{1}{c|}{} & \textbf{Abusive} & \textbf{Aggressive} & \textbf{Hate} & \textbf{Identity Hate} & \textbf{Insult} & \textbf{Neither} & \textbf{Obscene} & \textbf{Offensive} & \textbf{Profane} & \textbf{Severe Toxic} &\textbf{Threat} & \textbf{Toxic} \\
     \hline
     $\mathbf{G^{SUD}}$& 79.4 & 64.1 & 65.8 & 35.9 & 50 & 94.3 & 25.6 & 74.9 & 30.5 & 39.5 & 42.6 & 17.7 \\
     \hline
     \textbf{Davidson}   & - & - & 41.4 & - & - & 88.5 & - & 89.2 & - & - & - & - \\
     \hline
     \textbf{Founta} & 81.7 & - & 33.2 & - & - & 95.5 & - & - & - & - & - & - \\
     \hline
     \textbf{Fox}        & - & - & 13 & - & - & 82.6 & - & - & - & - & - & - \\
     \hline
     \textbf{Gab}        & - & - & 86.4 & - & - & 88.6 & - & - & - & - & - & - \\
     \hline
     \textbf{Grimminger} & - & - & 10.8 & - & - & 93 & - & - & - & - & - & - \\
     \hline
     \textbf{HASOC2019}  & - & - & 7.94 & - & - & 78.1 & - & 25 &  20.4 & - & - & - \\
     \hline
     \textbf{HASOC2020}  & - & - & 6.67 & - & - & 91.1 & - & 29.7 & 39.1 & - & - & - \\
     \hline
     \textbf{Hateval}    & - & - & 53.2 & - & - & 73.9 & - & - & - & - & - & - \\
     \hline
     \textbf{Jigsaw}     & - & - & - & 37.9 & 53.1 & 97.5 & 26.9 & - & - & 40.4 & 46 & 18.1 \\
     \hline
     \textbf{Olid}       & - & - & - & - & - & 85.8 & - & 45.3 & - & - & - & - \\
     \hline
     \textbf{Reddit}     & - & - & 74 & - & - & 89.5 & - & - & - & - & - & - \\
     \hline
     \textbf{Stormfront} & - & - & 39.7 & - & - & 94.1 & - & - & - & - & - & - \\
     \hline
     \textbf{Trac}       & - & 68.1 & - & - & - & 66.1 & - & - & - & - & - & - \\
     \hline
    \end{tabular}
}
\caption{Macro F1 Score of SUD classification per class and dataset.}
\label{tabPerClassPerDataset}
\end{table*}

Here, we observe that some class features, i.e., \emph{Abusive} (top-right), \emph{Aggressive} (bottom-center) form fairly clear clusters.
We can expect this behavior as each one of these class labels solely occurs in a single dataset, as depicted in Table~\ref{tabPerClassPerDataset}.  

Some other classes, i.e., \emph{Hate}, \emph{Offensive}, and \emph{Toxic}, have more sparse values, which is one reason behind the absolutely low F1 score.
Once again, these results get confirmed by the absolute low Macro F1 score both in the global corpus and in each single dataset.

Overall, the results explains the poor generalization capabilities of the studied classification model as this latter attains a low Macro F1 ($58\%$) score on $G^{SUD}$.  
In detail, we note that problematic classes are not only those with the lowest number of training samples as one might expect. 
In fact, a performance drop occur in $G^{SUD}$ classes that share samples from multiple corpus, suggesting the presence of intraclass heterogeneous samples as depicted in Table~\ref{tabPerClassPerDataset}.

In this sense, a clear example concerns the \emph{hate} class that contains samples from ten different datasets (out of thirteen). 
We note that shaky classification performance in each dataset of $G^{SUD}$ (see Table~\ref{tabPerClassPerDataset}) depends on divergent annotation criteria on a sensibly general concept, which can relate to different textual elements.

In Table~\ref{tabMultiClassBinary}, we depict the classification results obtained for each dataset in the global corpus $G^{SUD}$, and when the model was trained only using a single dataset (Individual).
We note that only in two cases the global model performs better than the individual counterpart (for the Fox and Grimminger datasets).
We believe that the relatively small support of these two corpora is the reason behind this improvement.
Nevertheless, leveraging more knowledge from multiple domains does not constitute an advantage in practice.

\begin{table}[htbp]
    \centering
\resizebox{8cm}{!}
{

    \begin{tabular}{ | c | c | c | c | c | }
     \cline{2-5}
     \multicolumn{1}{c|}{} & \multicolumn{4}{c|}{\textbf{Macro F1 Score (\%)}} \\
     \cline{2-5}
     \multicolumn{1}{c|}{} & \multicolumn{2}{|c|}{\textbf{(a) Multiclass SUD Classification}} && \multicolumn{1}{c|}{\textbf{(b) Binary Classification}} \\
    \hline
     \textbf{Dataset} &  \textbf{Classified in $\mathbf{G^{SUD}}$}  & \textbf{Individual}  && \textbf{Classified in $\mathbf{G^{SUD}}$}  \\
     \hline
                          $G^{SUD}$&   53.9        &     -         && 88.5  \\
     \hline
                          Davidson   &   73          & {75.1} && 93.9  \\ 
     \hline
                          Founta     &   70.1        & {74.7} && 92.9  \\    
     \hline
      Fox        & \textbf{47.8} &     41.6      && {59.2}  \\    
     \hline
      Gab        &   87.5        & {89.9} && 86.2  \\   
     \hline
      Grimminger & \textbf{51.9} &     46.9      && {64}    \\  
     \hline
                          HASOC2019  &   32.9        & {40.8} && 64.5  \\  
     \hline
                          HASOC2020  &   41.7        & {48.4} && 88.2  \\    
     \hline
      Hateval    &   63.6        & {75.7} && 70.2  \\
     \hline
                          Jigsaw     &   45.7        & {52.6} && 87.7  \\
     \hline
      Olid       &   65.6        & {75.2} && 72.3  \\    
     \hline
      Reddit     &   81.7        & {82.9} && 79.9  \\  
     \hline
      Stormfront &   66.9        & {76.1} && 71.1  \\   
     \hline
      Trac       &   67.1        & {73.1} && 69.3  \\    
     \hline
    \end{tabular}

}

\caption{(a) \textbf{Multiclass} SUD classification results (F1 score) with the model trained in $G^{SUD}$ VS on each single dataset. (a) \textbf{Binary} SUD classification with the model trained in $G^{SUD}$.}
\label{tabMultiClassBinary}
\end{table}

\subsection{Binary SUD Classification} \label{binaryClass}

For each of the experiments reported in this section, we have also tested the capability of the model to discriminate SUD and non-SUD text in $G^{SUD}$ irrespective of the specific class. 
To that extent, we use the same configuration for the classification head, changing the output layer to perform binary classification and re-training the model. 
For this case, we obtain a relatively high Macro F1 score ($\sim90\%$).
Such results suggest how the model discriminates well the \emph{neither} class from the generic SUD in the global context we built, confirming the current trend observed in the ML literature so far.
At the same time, effectively modeling multi-class SUD remains an open challenge.  

\section{Further Discussion and Perspective}

\begin{figure}[tb]
  \centering
  \includegraphics[trim={1cm 8cm 0cm 0cm}, scale=0.61]{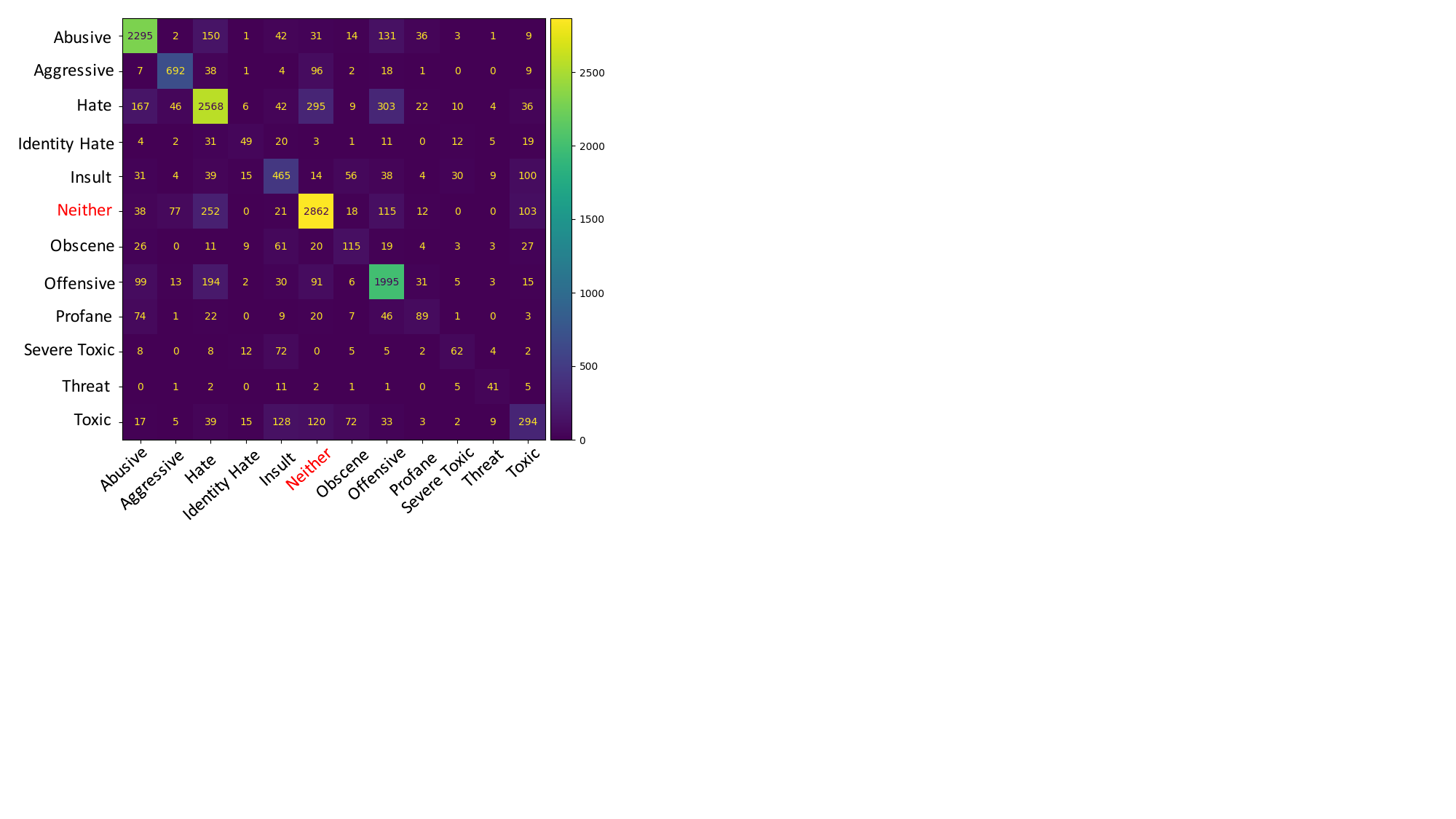}
  \caption{Confusion matrix of multi-class SUD classification.}
  \label{fig:confMatrix}
\end{figure}

To closely analyze the state-of-the-art limitation on SUD modeling, in figure~\ref{fig:confMatrix}, we plot the confusion matrix computed on the test set. 
In this case, we consider a test corpus with undersampled instances of neither class since, for this case, the classification model performs (slightly) in the best manner.
Here, we can observe multiple critical cases that concern the labels \emph{Identity Hate}, \emph{Toxic}, \emph{Obscene} and \emph{Profane}.
The classification model assigns a random label to these four classes that have overlapping features with all the others.
Concerning classification performance, we note that the F1 score is not significantly dropping for these classes when the model applies to $G^{SUD}$. It derives that learned features are fairly conserved in the new global context.

This observation confirms the results proposed by prior studies~\cite{yuan-etal-2022-separating,Fortuna2020ToxicHO}, which already analyzed the relation among several classes in significantly smaller corpora.

We believe the large-scale scenario we propose motivates the need for a more consistent effort in the ML community to equip language models with more discriminant power. 
This concerns the capability to distinguish the source and the target of the SUD discourse (individual rather than group), as well as the elements that characterize the kind of narrative of each SUD class.

\section{Conclusion and Future Work}

In this work, we present an empirical evaluation of automatic SUD detection using the BERT model, a state-of-the-art Deep Learning architecture for SUD classification.   
To test generalization capability, we consider a large and heterogeneous context in which we obtain results that are not in line with the expected performance of the model trained at the local level, i.e., in every single corpus.
In this sense, we argue that to build more general and reliable models, the ML community should consider formal guidelines provided by language experts (mostly neglected so far), which can sensibly reduce local bias (e.g., annotation policy, context, etc.). 
In future work, we plan to closely analyze the inter-domain mismatches we observe at the class sample level. 
Such effort would be beneficial to understand how to improve textual feature learning and to communicate requirements and expectations from the annotation task.

We furthermore note that the results and the insights we obtained also have the potential for the research linguists, discourse analysis, or semantics, as they show, from a knowledge base constituted by the main works on SUD corpora, the semantic links, and conceptual relationships, between several labels or tags.

In fact, over and above terminology, it is crucial to clearly state and understand the specific features of hate speech, offensive speech, or extremist speech. These initial results are necessary to foster several research discussions in the Horizon Europe ARENAS project into which this work integrates.

Specifically, the semantic issues in discourse categorization have an impact not only in terminological and computational terms (for annotating and classifying) but also in legal, political, and sociological terms. 
The impact of different characterizations is not neutral, there are potential issues of moderation or condemnation~\cite{LONGHI2021110564}, and it is necessary to proceed cautiously and rigorously in the delimitation of the chosen descriptors and in the way they are defined and characterized.

Finally,  the explicability of these categories and the classification provided by Artificial Intelligence is central to future research. 
Making transparent outcomes will enable us to propose valuable results for all those involved in the hate speech and extremism analysis. 
In the context of a multidisciplinary project like ARENAS, which brings together scientists with different backgrounds (i.e., linguists, political scientists, etc.) and targets a heterogeneous audience, such as lawyers and journalists, the clarity of descriptors, and their ability to be understood by different stakeholders, is an essential element.

\section{References}

\bibliographystyle{lrec2016}
\bibliography{bibliography.bib}

\begin{thebibliography}{}

\bibitem[\protect\citename{Aroyehun and Gelbukh}2018]{trac_w}
Aroyehun, S.~T. and Gelbukh, A.
\newblock (2018).
\newblock Aggression detection in social media: Using deep neural networks,
  data augmentation, and pseudo labeling.
\newblock In {\em {TRAC}-2018}.

\bibitem[\protect\citename{Ascone and Longhi}2018]{Ascone_Longhi_2018}
Ascone, L. and Longhi, J.
\newblock (2018).
\newblock The expression of threat in jihadist propaganda.
\newblock {\em Fragmentum}.

\bibitem[\protect\citename{Badjatiya \bgroup et al.\egroup }2019]{bias1}
Badjatiya, P., Gupta, M., and Varma, V.
\newblock (2019).
\newblock Stereotypical bias removal for hate speech detection task using
  knowledge-based generalizations.
\newblock In {\em The World Wide Web Conference}.

\bibitem[\protect\citename{Basile \bgroup et al.\egroup }2019]{hateval}
Basile, V., Bosco, C., Fersini, E., Nozza, D., Patti, V., Rangel~Pardo, F.~M.,
  Rosso, P., and Sanguinetti, M.
\newblock (2019).
\newblock {S}em{E}val-2019 task 5: Multilingual detection of hate speech
  against immigrants and women in {T}witter.
\newblock In {\em Proceedings of the 13th International Workshop on Semantic
  Evaluation}, June.

\bibitem[\protect\citename{Conneau \bgroup et al.\egroup
  }2020]{ConneauKGCWGGOZ20}
Conneau, A., Khandelwal, K., Goyal, N., Chaudhary, V., Wenzek, G.,
  Guzm{\'{a}}n, F., Grave, E., Ott, M., Zettlemoyer, L., and Stoyanov, V.
\newblock (2020).
\newblock Unsupervised cross-lingual representation learning at scale.
\newblock In {\em {ACL} 2020, Online, July 5-10, 2020}.

\bibitem[\protect\citename{Davidson \bgroup et al.\egroup }2017]{davidson}
Davidson, T., Warmsley, D., Macy, M., and Weber, I.
\newblock (2017).
\newblock Automated hate speech detection and the problem of offensive
  language.

\bibitem[\protect\citename{Davidson \bgroup et al.\egroup }2019]{racial_bias}
Davidson, T., Bhattacharya, D., and Weber, I.
\newblock (2019).
\newblock Racial bias in hate speech and abusive language detection datasets.
\newblock In {\em Proceedings of the Third Workshop on Abusive Language
  Online}.

\bibitem[\protect\citename{de Gibert \bgroup et al.\egroup }2018]{stormfront}
de~Gibert, O., Perez, N., Garc{\'\i}a-Pablos, A., and Cuadros, M.
\newblock (2018).
\newblock {Hate Speech Dataset from a White Supremacy Forum}.
\newblock In {\em {ALW}2}, October.

\bibitem[\protect\citename{De~Giorgio \bgroup et al.\egroup
  }2022]{vaccines10030481}
De~Giorgio, A., Kuva\v{s}i{\'c}, G., Male\v{s}, D., Vecchio, I., Tornali, C.,
  Ishac, W., Ramaci, T., Barattucci, M., and Milavi{\'c}, B.
\newblock (2022).
\newblock Willingness to receive covid-19 booster vaccine: Associations between
  green-pass, social media information, anti-vax beliefs, and emotional
  balance.
\newblock {\em Vaccines}, 10.

\bibitem[\protect\citename{de Maiti \bgroup et al.\egroup
  }2020]{Maiti2020GrammaticalFO}
de~Maiti, K.~P., Fi{\v{s}}er, D., and Erjavec, T.
\newblock (2020).
\newblock Grammatical footprint of socially unacceptable facebook comments.
\newblock In {\em Language Technologies \& Digital Humanities}.

\bibitem[\protect\citename{Devlin \bgroup et al.\egroup }2019]{bert}
Devlin, J., Chang, M.-W., Lee, K., and Toutanova, K.
\newblock (2019).
\newblock Bert: Pre-training of deep bidirectional transformers for language
  understanding.

\bibitem[\protect\citename{Fi{\v{s}}er \bgroup et al.\egroup
  }2017]{fiser-etal-2017-legal}
Fi{\v{s}}er, D., Erjavec, T., and Ljube{\v{s}}i{\'c}, N.
\newblock (2017).
\newblock Legal framework, dataset and annotation schema for socially
  unacceptable online discourse practices in {S}lovene.
\newblock In {\em Proceedings of the First Workshop on Abusive Language
  Online}.

\bibitem[\protect\citename{Fortuna \bgroup et al.\egroup
  }2020]{Fortuna2020ToxicHO}
Fortuna, P., Soler, J., and Wanner, L.
\newblock (2020).
\newblock Toxic, hateful, offensive or abusive? what are we really classifying?
  an empirical analysis of hate speech datasets.
\newblock In {\em International Conference on Language Resources and
  Evaluation}.

\bibitem[\protect\citename{Founta \bgroup et al.\egroup }2018]{founta}
Founta, A.-M., Djouvas, C., Chatzakou, D., Leontiadis, I., Blackburn, J.,
  Stringhini, G., Vakali, A., Sirivianos, M., and Kourtellis, N.
\newblock (2018).
\newblock Large scale crowdsourcing and characterization of twitter abusive
  behavior.

\bibitem[\protect\citename{Gao and Huang}2018]{fox}
Gao, L. and Huang, R.
\newblock (2018).
\newblock Detecting online hate speech using context aware models.

\bibitem[\protect\citename{Grimminger and Klinger}2021]{grimminger}
Grimminger, L. and Klinger, R.
\newblock (2021).
\newblock Hate towards the political opponent: A {T}witter corpus study of the
  2020 {US} elections on the basis of offensive speech and stance detection.
\newblock In {\em Proceedings of the Eleventh Workshop on Computational
  Approaches to Subjectivity, Sentiment and Social Media Analysis}, April.

\bibitem[\protect\citename{Jasser \bgroup et al.\egroup }2021]{gab_info}
Jasser, G., McSwiney, J., Pertwee, E., and Zannettou, S.
\newblock (2021).
\newblock Welcome to \#gabfam: Far-right virtual community on gab.
\newblock {\em New Media \& Society}.

\bibitem[\protect\citename{Kumar \bgroup et al.\egroup }2018]{trac}
Kumar, R., Reganti, A.~N., Bhatia, A., and Maheshwari, T.
\newblock (2018).
\newblock {Aggression-annotated Corpus of Hindi-English Code-mixed Data}.
\newblock In {\em (LREC 2018)}.

\bibitem[\protect\citename{Longhi}2021]{LONGHI2021110564}
Longhi, J.
\newblock (2021).
\newblock Using digital humanities and linguistics to help with terrorism
  investigations.
\newblock {\em Forensic Science International}, 318:110564.

\bibitem[\protect\citename{MacAvaney \bgroup et al.\egroup }2019]{mSVM}
MacAvaney, S., Yao, H.-R., Yang, E., Russell, K., Goharian, N., and Frieder, O.
\newblock (2019).
\newblock Hate speech detection: Challenges and solutions.
\newblock {\em PLOS ONE}.

\bibitem[\protect\citename{Machado~Carneiro \bgroup et al.\egroup
  }2023]{paperRepo}
Machado~Carneiro, B., Linardi, M., and Longhi, J.
\newblock (2023).
\newblock \url{https://github.com/mlinardiCYU/SUD\_study\_different\_eyes.git}.

\bibitem[\protect\citename{Mandl \bgroup et al.\egroup }2020]{hasoc2020}
Mandl, T., Modhab, S., Shahic, G.~K., Jaiswald, A.~K., Nandinie, D., Patelf,
  D., Majumderg, P., and Sch\"{a}fera, J.
\newblock (2020).
\newblock Overview of the hasoc track at fire 2020: Hate speech and offensive
  content identification in indo-european languages.

\bibitem[\protect\citename{Modha \bgroup et al.\egroup }2019]{hasoc2019}
Modha, S., Mandl, T., Majumder, P., and Pate, D.
\newblock (2019).
\newblock Overview of the hasoc track at fire 2019: Hate speech and offensive
  content identification in indo-european languages.

\bibitem[\protect\citename{Qian \bgroup et al.\egroup }2019]{reddit_gab}
Qian, J., Bethke, A., Liu, Y., Belding, E., and Wang, W.~Y.
\newblock (2019).
\newblock A benchmark dataset for learning to intervene in online hate speech.

\bibitem[\protect\citename{Roy \bgroup et al.\egroup }2021]{hasoc2020_w}
Roy, S.~G., Narayan, U., Raha, T., Abid, Z., and Varma, V.
\newblock (2021).
\newblock Leveraging multilingual transformers for hate speech detection.

\bibitem[\protect\citename{Sulc and de Maiti}2020]{Sulc2020NoRF}
Sulc, A. and de~Maiti, K.~P.
\newblock (2020).
\newblock No room for hate: What research about hate speech taught us about
  collaboration?
\newblock In {\em TwinTalks@DH/DHN}.

\bibitem[\protect\citename{Swamy \bgroup et al.\egroup }2019]{swamy}
Swamy, S.~D., Jamatia, A., and Gamb{\"a}ck, B.
\newblock (2019).
\newblock Studying generalisability across abusive language detection datasets.
\newblock In {\em Proceedings of the 23rd Conference on Computational Natural
  Language Learning (CoNLL)}.

\bibitem[\protect\citename{Toraman \bgroup et al.\egroup
  }2022]{toraman2022large}
Toraman, C., \c{S}ahinu\c{c}, F., and Yilmaz, E.~H.
\newblock (2022).
\newblock Large-scale hate speech detection with cross-domain transfer.
\newblock In {\em Proceedings of the Language Resources and Evaluation
  Conference}.

\bibitem[\protect\citename{van Aken \bgroup et al.\egroup }2018]{jigsaw_w}
van Aken, B., Risch, J., Krestel, R., and L\"{o}ser, A.
\newblock (2018).
\newblock Challenges for toxic comment classification: An in-depth error
  analysis.

\bibitem[\protect\citename{Wang \bgroup et al.\egroup }2019]{hasoc19_winner}
Wang, B., Ding, Y., Liu, S., and Zhou, X.
\newblock (2019).
\newblock Ynu\_wb at hasoc 2019: Ordered neurons lstm with attention for
  identifying hate speech and offensive language.
\newblock In {\em Fire}.

\bibitem[\protect\citename{Yuan and Rizoiu}2022]{yuan_unseen}
Yuan, L. and Rizoiu, M.-A.
\newblock (2022).
\newblock Detect hate speech in unseen domains using multi-task learning: A
  case study of political public figures.

\bibitem[\protect\citename{Yuan \bgroup et al.\egroup }2022a]{yuan_transfer}
Yuan, L., Wang, T., Ferraro, G., Suominen, H., and Rizoiu, M.-A.
\newblock (2022a).
\newblock Transfer learning for hate speech detection in social media.

\bibitem[\protect\citename{Yuan \bgroup et al.\egroup
  }2022b]{yuan-etal-2022-separating}
Yuan, S., Maronikolakis, A., and Sch{\"u}tze, H.
\newblock (2022b).
\newblock Separating hate speech and offensive language classes via adversarial
  debiasing.
\newblock In {\em Proceedings of the Sixth Workshop on Online Abuse and Harms
  (WOAH)}.

\bibitem[\protect\citename{Zampieri \bgroup et al.\egroup }2019]{olid}
Zampieri, M., Malmasi, S., Nakov, P., Rosenthal, S., Farra, N., and Kumar, R.
\newblock (2019).
\newblock Predicting the type and target of offensive posts in social media.
\newblock In {\em Proceedings of the 2019 Conference of the North {A}merican
  Chapter of the Association for Computational Linguistics: Human Language
  Technologies, Volume 1 (Long and Short Papers)}.

\end{thebibliography}

\end{document}